\documentclass[sigconf]{acmart}
\AtBeginDocument{%
  \providecommand\BibTeX{{%
    \normalfont B\kern-0.5em{\scshape i\kern-0.25em b}\kern-0.8em\TeX}}}

\setcopyright{acmcopyright}
\copyrightyear{2020}
\acmYear{2020}
\acmDOI{10.1145/1122445.1122456}

\acmConference[DLP-KDD'20]{2020 Association for Computing Machinery}{August 24, 2020}{San Diego, California USA}
\acmPrice{15.00}
\acmISBN{978-1-4503-6317-4/19/07}

\begin{document}

\title{Correct Normalization Matters: Understanding the Effect of Normalization On Deep Neural Network Models For Click-Through Rate Prediction}

\author{Zhiqiang Wang, Qingyun She, PengTao Zhang, Junlin Zhang}
\affiliation{
     \institution{Sina Weibo Corp}
     \city{Beijing}
     \country{China}}
\email{{zhiqiang36,qingyun,pengtao1,junlin6}@staff.weibo.com}



\renewcommand{\shortauthors}{}

\begin{abstract}
  Normalization has become one of the most fundamental components in many deep neural networks for machine learning tasks while deep neural network has also been widely used in CTR estimation field. Among most of the proposed deep neural network models, few model utilize normalization approaches. Though some works such as Deep \& Cross Network (DCN) and Neural Factorization Machine (NFM) use Batch Normalization in MLP part of the structure, there isn't work to thoroughly explore the effect of the normalization on the DNN ranking systems. In this paper, we conduct a systematic study on the effect of widely used normalization schemas by applying the various normalization approaches to both feature embedding and MLP part in DNN model. Extensive experiments are conduct on three real-world datasets and the experiment results demonstrate that the correct normalization significantly enhances model's performance. We also propose a new and effective normalization approaches based on LayerNorm named variance only LayerNorm(VO-LN) in this work. A normalization enhanced DNN model named NormDNN is also proposed based on the above-mentioned observation. As for the reason why normalization works for DNN models in CTR estimation, we find that the variance of normalization plays the main role and give an explanation in this work.
\end{abstract}




\maketitle

\section{Introduction}
Normalization has become one of the most fundamental components in many deep neural networks for machine learning tasks, especially in Computer Vision (CV) and Natural Language Processing (NLP). However, very different kinds of normalization are used in CV and NLP. For example, Batch Normalization (BatchNorm or BN) \cite{ioffe2015batch} is widely adopted in CV, but it leads to significant performance degradation when naively used in NLP. Instead, Layer Normalization (LayerNorm or LN) \cite{ba2016layer} is the standard normalization method utilized in NLP.

On the other side, deep neural network has also been widely used in CTR estimation field \cite{10.1145/2487575.2488200,zhou2018deep,he2014practical,xiao2017attentional,qu2016product,he2017neural,zhang2016deep,cheng2016wide,guo2017deepfm,lian2018xdeepfm,beutel2018latent,huang2019fibinet,zhang2019fat}. Some deep learning based models have been introduced and achieved success such as wide \& deep \cite{cheng2016wide}, DeepFM\cite{guo2017deepfm} and xDeepFM\cite{lian2018xdeepfm} .Most DNN ranking models  use the feature embedding to represent information and shallow MLP layers to model high-order interactions in an implicit way. These two commonly used components play important roles in current state-of-the-art ranking systems.

Taking the both sides into consideration, we care about the following questions: What's the effect of various normalization methods on deep neural network models for CTR estimation? Is there one method outperforming other normalization approaches just like LayerNorm in NLP or BatchNorm in CV? What's the reason if some normalization works?

Among most of the proposed deep neural network models, few of them utilize normalization approaches. Though some  works such as DCN\cite{wang2017deep} and Neural Factorization Machine (NFM) \cite{he2017neural} use BatchNorm in MLP part of the structure,  there isn't work  to thoroughly explore the effect of the normalization on the DNN ranking systems. In this paper, we conduct a systematic study on the effect of widely used normalization schemas by applying the various normalization approaches to both feature embedding and MLP part in DNN model.  Experimental results show the correct normalization helps the training of DNN models and boosts the model performance with a large margin. We also simplify the LayerNorm and propose a new and effective normalization method in this work. A normalization enhanced DNN model named NormDNN is also proposed based on the above-mentioned observation. Further more, we find the variance of normalization mainly contributes to this positive effect. To the best of our knowledge, this is the first work to verify the importance of normalization on DNN ranking system through systematic study.

The contributions of our work are summarized as follows:
\begin{enumerate}

    \item	In this work, we propose a new normalization approach based on LayerNorm: variance-only LayerNorm(VO-LN). The experimental results show that the proposed normalization method has comparable performance with layer normalization and significantly enhance DNN model's performance.

    \item We apply various normalization approaches to the feature embedding part and the MLP part of DNN model, including commonly used normalization and our proposed approach. Extensive experiments are conduct on three real-world datasets and the experiment results demonstrate that the correct normalization or normalization combination significantly enhances model's performance. As far as we know, this is the first work to apply normalization to feature embedding and prove its effectiveness.

    \item We propose NormDNN model in this paper which is a normalization enhanced DNN adopting the following normalization strategy: variance-only LayerNorm or LayerNorm for numerical feature, BatchNorm for categorical feature and variance-only LayerNorm for MLP. NormDNN achieves significantly better performance than complex model such as xDeepFM. NormDNN is more applicable in many  industry applications because of its better performance and high computation efficiency compared with many state-of-the-art complex neural network models.

    \item To prove the universal validity of normalization for neural network ranking model, we also apply several normalization approaches to DeepFM and xDeepFM model. The experiments results imply that the correct normalization also boosts these model's performances with a large margin.

    \item As for the reason why normalization works for DNN models in CTR estimation, we find that the variance of normalization plays the main role and give an explanation in this paper.
\end{enumerate}

The rest of this paper is organized as follows. Section 2 introduces some related works which are relevant with our work. We introduce our proposed models in detail in Section 3. The experimental results on three real world datasets are presented and discussed in Section 4. Section 5 concludes our work in this paper.

\section{Related Work}
\subsection{Normalization}
Normalization techniques have been recognized as very effective components in deep learning. Many normalization approaches have been proposed with the three most popular ones being BatchNorm\cite{ioffe2015batch}, LayerNorm \cite{ba2016layer} and GroupNorm\cite{wu2018group}. Batch Normalization (Batch Norm or BN)\cite{ioffe2015batch} normalizes the features by the mean and variance computed within a mini-batch. This has been shown by many practices to ease optimization and enable very deep networks to converge. Another example is layer normalization (Layer Norm or LN)\cite{ba2016layer} which was proposed to ease optimization of recurrent neural networks. Statistics of layer normalization are not computed across the N samples in a mini-batch but are estimated in a layer-wise manner for each sample independently. It's an easy way to  extend LayerNorm to GroupNorm (GN)\cite{wu2018group}, where the normalization is performed across a partition of the features/channels with different pre-defined groups. Normalization methods have shown success in accelerating the training of deep networks. In general, BatchNorm \cite{ioffe2015batch} and GroupNorm \cite{wu2018group} are widely adopted in CV and LayerNorm  \cite{ba2016layer} is the standard normalization scheme used in NLP.

Another line of research on normalization is to understand why BatchNorm helps training in CV and LayerNorm helps training in NLP. For example, the original explanation was that BatchNorm reduces the so-called "Internal Covariance Shift" \cite{ioffe2015batch}. However, this explanation was viewed as incorrect or incomplete and the study of \cite{santurkar2018does} argued that the underlying reason that BatchNorm helps training is that it results in a smoother loss landscape. Shen etc \cite{shen2020rethinking} perform a systematic study of NLP transformer models to understand why BatchNorm has a poor performance and find that the statistics of NLP data across the batch dimension exhibit large fluctuations throughout training which results in instability. Xu etc \cite{xu2019understanding} find that the derivatives of the mean and variance in LayerNorm used in Transformer for NLP tasks are more important than forward normalization by re-centering and re-scaling backward gradients. Furthermore, they find that the parameters of LayerNorm, including the bias and gain, increase the risk of over-fitting and do not work in most cases.

\subsection{Neural Network based CTR Models}
Many deep learning based CTR models have been proposed in recent years and it is the key factor for most of these neural network based models to effectively model the feature interactions or feature importance.

Factorization-Machine Supported Neural Networks (FNN)\cite{zhang2016deep} is a feed-forward neural network using FM to pre-train the embedding layer. Wide \& Deep Learning\cite{cheng2016wide} jointly trains wide linear models and deep neural networks to combine the benefits of memorization and generalization for recommender systems. However, expertise feature engineering is still needed on the input to the wide part of Wide \& Deep model. To alleviate manual efforts in feature engineering, DeepFM\cite{guo2017deepfm} replaces the wide part of Wide \& Deep model with FM and shares the feature embedding between the FM and deep component. FiBiNet\cite{huang2019fibinet} and FAT-DeepFFM\cite{zhang2019fat} dynamically learn the importance of features via the Squeeze-Excitation network (SENET) mechanism based on different backbone networks.

While most DNN ranking models process high-order feature interactions in implicit way, some works explicitly introduce high-order feature interactions by sub-network. Deep \& Cross Network (DCN)\cite{wang2017deep} efficiently captures feature interactions of bounded degrees in an explicit fashion. Similarly, eXtreme Deep Factorization Machine (xDeepFM) \cite{lian2018xdeepfm} also models the low-order and high-order feature interactions in an explicit way by proposing a novel Compressed Interaction Network (CIN) part. AutoInt\cite{song2019autoint} proposes a multi-head self-attentive neural network with residual connections to explicitly model the feature interactions in the low-dimensional space.

\section{Our Work}
In this section, we first describe the proposed variance-only LayerNorm. We conduct extensive experiments to verify the effectiveness of normalization in section 4 and the details about how to apply the normalization on feature embedding and MLP will be introduced in this section. Finally the reason why normalization works is introduced.

\subsection{Variance-Only LayerNorm}
First, we briefly review the formulation of LayerNorm. Let $\mathbf{x} = (x_1, x_2,...,x_H)$ denotes the vector representation of an input of size $H$ to normalization layers. LayerNorm re-centers and re-scales input $\mathbf{x}$ as

\begin{equation}
  \begin{split}
  &\mathbf{h} = \mathbf{g} \odot N(\mathbf{x}) + \mathbf{b}, \quad N(\mathbf{x}) = \frac{\mathbf{x}-\mu}{\delta}, \\
  & \mu = \frac{1}{H}\sum^H_{i=1}x_i, \quad \delta = \sqrt{\frac{1}{H}\sum^{H}_{i=1}(x_i - \mu)^2}
\end{split}
\end{equation}
where $\mathbf{h}$ is the output of a LayerNorm layer. $\odot$ is a dot production operation. $\mu$ and $\delta$ are the mean and standard deviation of input. Bias $\mathbf{b}$ and gain $\mathbf{g}$ are parameters with the same dimension $H$.

As we all know, LayerNorm has been widely used and proved to be effective in many NLP tasks. However, Xu etc.\cite{xu2019understanding} point out that the parameters of LayerNorm, including the bias and gain, increase the risk of over-fitting and do not work in most cases. Their experiments in four NLP datasets show that a simple version of LayerNorm without the bias and gain outperforms LayerNorm. Though their conclusion is draw mainly from the NLP tasks and they primarily consider normalization on Transformer and Transformer-XL networks. We wonder whether the same conclusion can be draw in DNN ranking models and design a similar simper version LayerNorm by removing bias and gain from LN as follows:
\begin{equation}
  \begin{split}
  &\mathbf{h} = N(\mathbf{x}) , \quad N(\mathbf{x}) = \frac{\mathbf{x}-\mu}{\delta}, \\
  &\mu = \frac{1}{H}\sum^H_{i=1}x_i, \quad \delta = \sqrt{\frac{1}{H}\sum^{H}_{i=1}(x_i - \mu)^2}
\end{split}
\end{equation}

We call this version LayerNorm simple-LayerNorm (S-LN) just as the original paper \cite{xu2019understanding} named. Our experimental results show that simple-LayerNorm has comparable performance with LayerNorm, which implies the bias and gain in LayerNorm bring neither good nor bad effect to DNN models in CTR estimation field. Our conclusion is slightly different from that in NLP field because their experimental results \cite{xu2019understanding} show the advantages for simple-LayerNorm over the standard LayerNorm in several NLP tasks. We deem that may come from the difference of network structure and research field.

According to our empirical observations, we find re-centering the input x in simple-LayerNorm has little effect on the performance of DNN ranking model. So we propose variance-only LayerNorm(VO-LN) by further removing the mean from simple-LayerNorm as follows:

\begin{equation}
  h = \frac{\mathbf{x}}{\delta}, \quad \mu = \frac{1}{H}\sum^H_{i=1}x_i, \quad \delta = \sqrt{\frac{1}{H}\sum^{H}_{i=1}(x_i - \mu)^2}
\end{equation}

Though the variance-only LayerNorm seems rather simple, our experimental results demonstrate it has comparable or even better performance in several CTR datasets than standard LayerNorm.

\subsection{NormDNN}
Most DNN ranking models use the feature embedding to represent information and shallow MLP layers to model high-order interactions in an implicit way. These two commonly used components play important roles in current state-of-the-art ranking systems. So we have three options to apply normalization: normalization only on feature embedding, normalization only on MLP part, normalization both on feature embedding and MLP part.

We also find different parts of DNN model need different normalization method and propose the following unified normalization combination strategy: variance-only LayerNorm or LayerNorm for numerical feature, BatchNorm for categorical feature and variance-only LayerNorm for MLP.  We call this normalization enhanced DNN model with this unified normalization strategy "NormDNN" in this paper. NormDNN achieves significantly better performance than complex model such as xDeepFM.  We will discuss this in Section 4.6.

 \subsubsection{\textbf{Normalization on Feature Embedding}}

The input data of CTR tasks usually consists of sparse and dense features and the sparse features are mostly categorical type. Such features are encoded as one-hot vectors which often lead to excessively high-dimensional feature spaces for large vocabularies. The common solution to this problem is to introduce the embedding layer.

An embedding layer is applied upon the raw feature input to compress it to a low dimensional, dense real-value vector. The result of embedding layer is a wide concatenated vector:

\begin{equation}
  V_{emb} = concat(e_1, e_2, ..., e_i, ..., e_f)
\end{equation}
where $f$ denotes the number of fields, and $e_i \in \mathbb{R}^{k}$ denotes the embedding of one field. Although the feature lengths of instances can be various, their embedding are of the same length $f \times k$, where $k$ is the dimension of field embedding.

As we all know, features in CTR tasks usually can be segregated into categorical features and numerical features. There are two widely used approaches to convert the numerical feature into embedding. The first one is to quantize each numerical feature into discrete buckets, and the feature is then represented by the bucket ID. We can map bucket ID to an embedding vector. The second method maps the feature field into an embedding vector as follows:

\begin{equation}
  \mathbf{v_i} = \mathbf{e}_i\mathbf{x}_i
\end{equation}
where $\mathbf{e}_i$ is an embedding vector for field $i$, and $\mathbf{x}_i$ is a scalar value. In our experiments, we adopt the second approach to convert numerical features into embedding.

 We apply normalization on feature embedding based on the feature field as follows:

\begin{equation}
  N(V_{emb}) = concat(N(e_1), N(e_2), ...,N(e_i),..., N(e_f))
\end{equation}
where $N$ can be BatchNorm, LayerNorm, GroupNorm, Simple-LayerNorm or variance-only LayerNorm. The bias and gain are shared for features in same feature field if the normalization contains these parameters.

For the LayerNorm based normalization approaches (LayerNorm, Simple-LayerNorm and variance-only LayerNorm), we regard each feature's embedding as a layer to compute the mean and variance of normalization. As for the GroupNorm, the feature embedding is divided into several groups to compute mean and variance. BatchNorm computes the statistics within a mini-batch.

In real life applications, the CTR tasks usually contain both categorical features and numerical features. We find the different kinds of feature need corresponding normalization method and we will discuss this in detail in Section 4.2.

 \subsubsection{\textbf{Normalization on MLP Part}}
As for the feed-forward layer in DNN model, the normalization on MLP is just as usual method does. That is to say, BatchNorm's  mean and variance are computed within a mini-batch and  LayerNorm based normalizations's statistics are estimated in a layer-wise manner. As for the GroupNorm, we can divide the neural units contained in MLP into several groups and the statistics are estimated in group-wise manner.

Notice that we have two places to put normalization operation on the MLP: one place is before non-linear operation and another place is after non-linear operation. For clarity of the description, we use LayerNorm as an example.  If we put normalization after non-linear operation, we have:
\begin{equation}
  LN(V_{hidden}) = LN(ReLU(W_i\mathbf{x}))
\end{equation}
where $\mathbf{x} \in \mathbb{R}^t$ refers to the input of feed-forward layer, $W_i \in \mathbb{R}^{m\times t}$ are parameters for the layer, $t$ and $m$ respectively denotes the  size of input layer and neural number of feed-forward layer.

If we put normalization before non-linear operation, we have:
\begin{equation}
  LN(V_{hidden}) = ReLU(LN(W_i\mathbf{x}))
\end{equation}

We find the performance of the normalization before non-linear consistently outperforms that of the normalization after non-linear operation. So all the normalization used in MLP part is put before non-linear operation in our paper.

\begin{figure}[!]
  \includegraphics[width=1.0\linewidth]{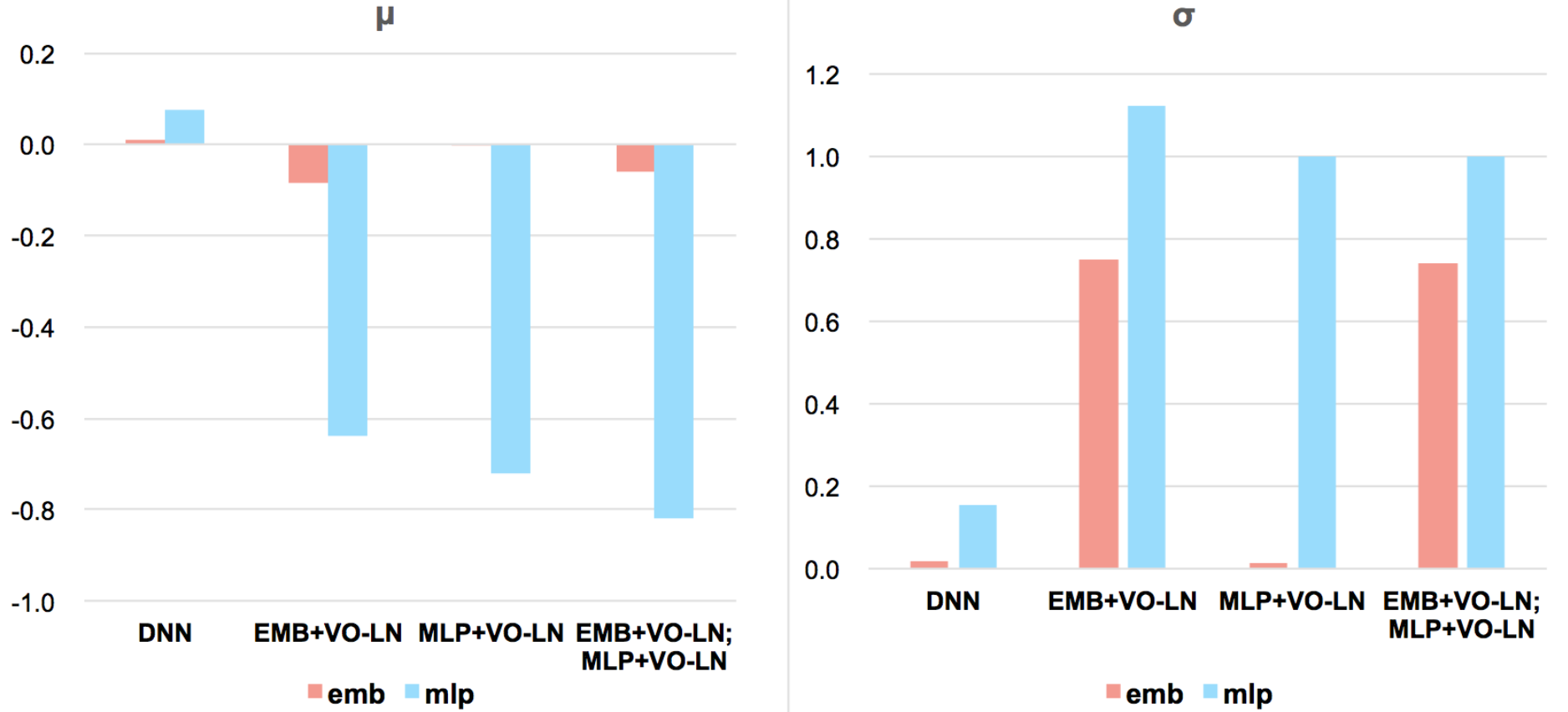}
  \caption{Average Statistics of Different Models}
  \label{fig:fig1}
\end{figure}

\begin{figure}[!]
  \includegraphics[width=1.0\linewidth]{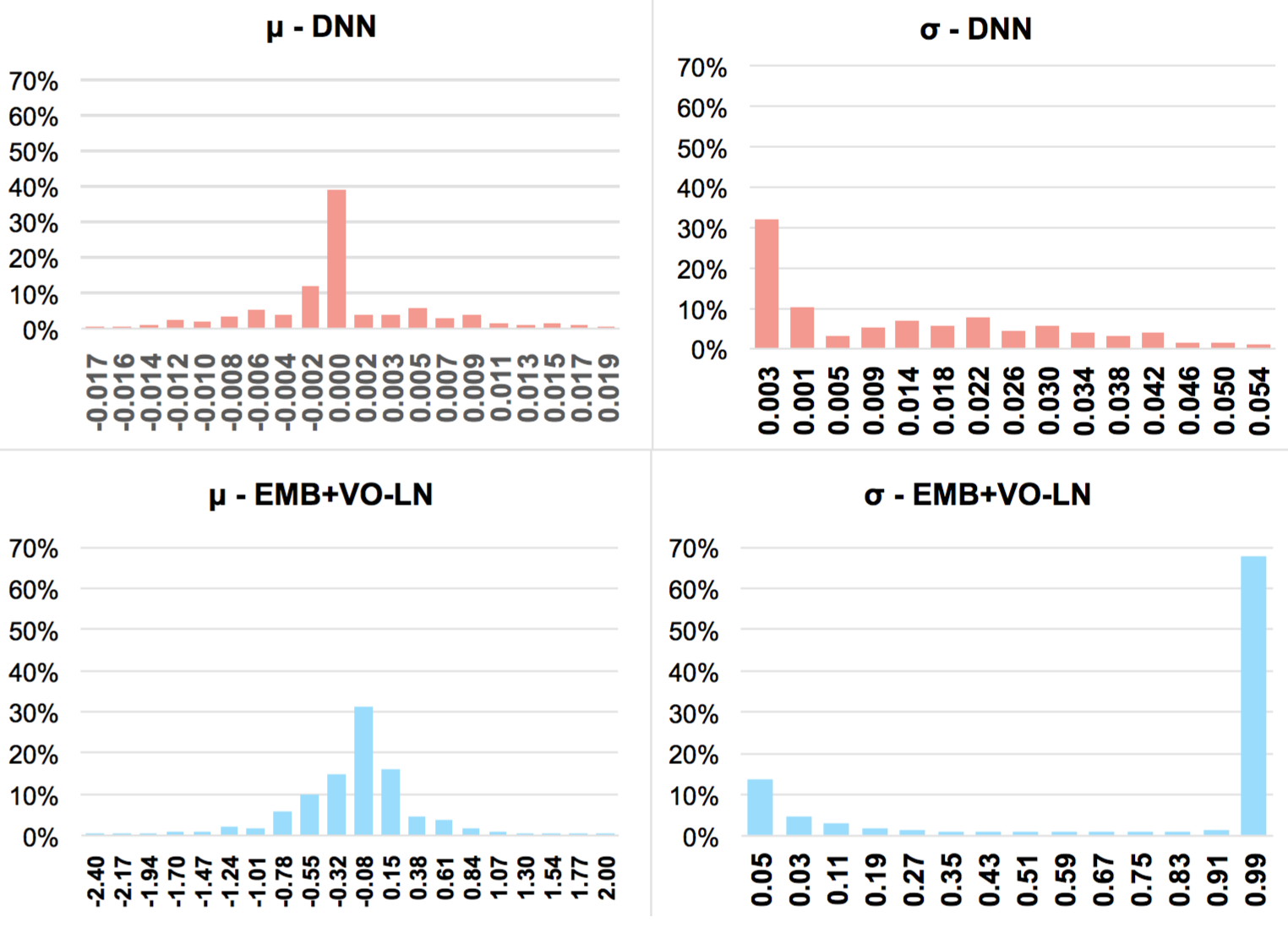}
  \caption{Statistics Changes after VO-LN on Feature Embedding}
  \label{fig:fig2}
\end{figure}

\subsection{Understanding Why Normalization Works}
In this section, we discuss the reason why normalization works in DNN model. The related experimental results are presented in Section 4.2.

As mentioned in Section 3.1, we find three LayerNorm based models( LayerNorm, Simple-LayerNorm and variance-only LayerNorm) have comparable performance on three real-life datasets and all the three normalization approaches work. From these observations, we can draw the following conclusion: Because the simple LayerNorm just removes the bias and gain from LayerNorm, the similar model performance implies the bias and gain have no effect on the final model performance. Further more, the variance-only LayerNorm removes the mean from simple LayerNorm and it has comparable performance with LayerNorm and simple LayerNorm. That implies the mean in simple LayerNorm doesn't contribute to the final better performance. So we can draw the conclusion that it's variance in variance-only LayerNorm that helps boosting model's performance.  The reason why LayerNorm and simple LayerNorm also work lies in that they contain variance.

\begin{figure}
  \includegraphics[width=1.0\linewidth]{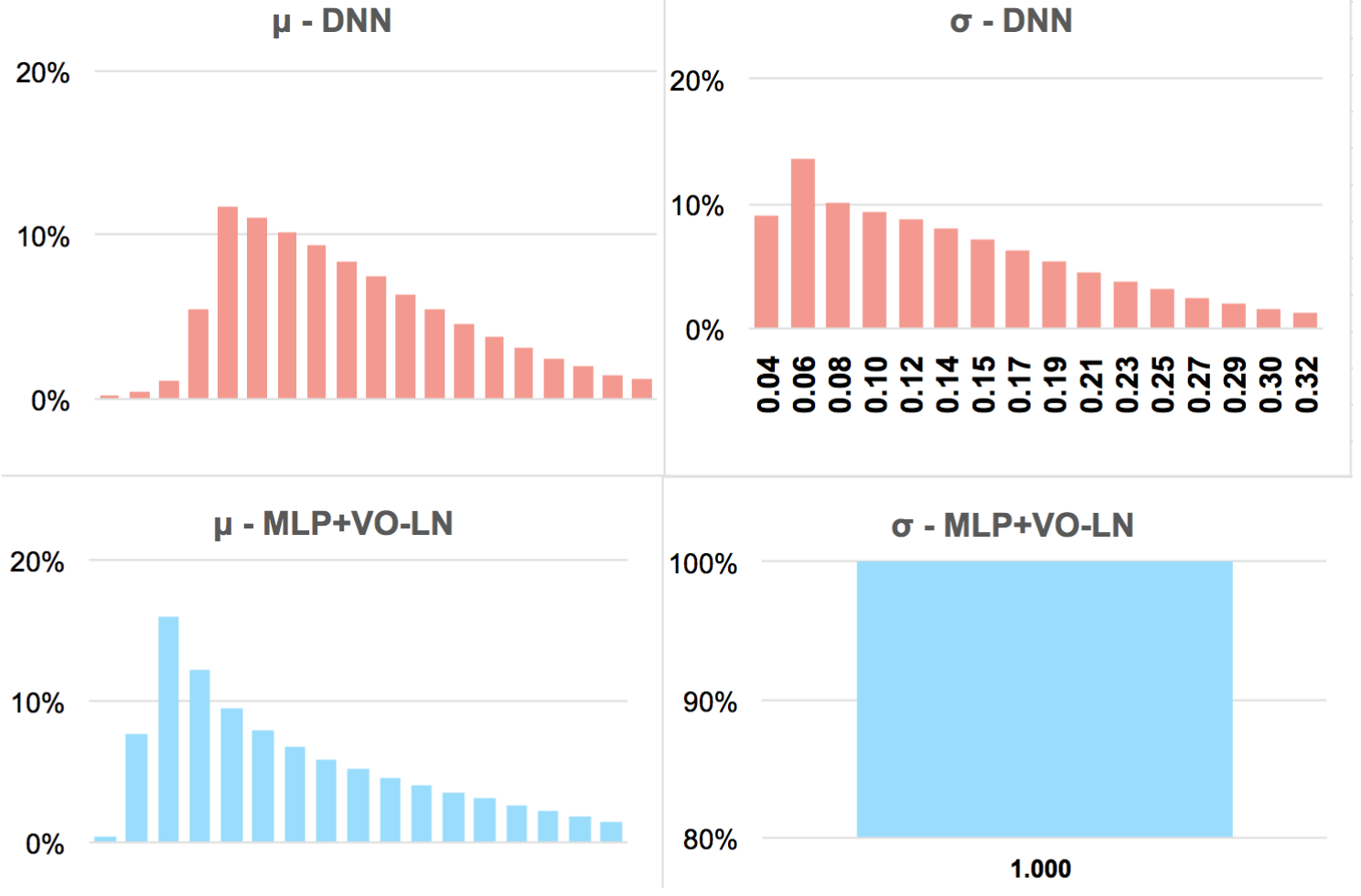}
  \caption{Statistics Changes after VO-LN on MLP Part}
  \label{fig:fig3}
\end{figure}

To understand how the variance influences the model performance, we analyze the change of statistics both in embedding layer and MLP after applying variance-only LayerNorm on Cretio dataset (Figure. \ref{fig:fig1}).  From Figure \ref{fig:fig1},we can see that the average mean and variance of feature embedding and MLP are very small positive number if we don't apply normalization on any part of DNN model. If we use variance-only LayerNorm only on feature embedding, the variance of feature embedding greatly increases and that change pushes bit value of many feature embedding to a much larger negative number (Figure \ref{fig:fig1} and Figure \ref{fig:fig2}). Through the network connections, these statistics changes of feature embedding are transferred to the MLP layer and the corresponding statistics of MLP show the similar trend even though we didn't apply any normalization on it (Figure \ref{fig:fig1}). If we just apply variance-only LayerNorm on MLP of DNN model, we see the similar changes that the average variance of MLP neurons greatly increases and that also pushes output of many neurons to negative number (Figure \ref{fig:fig3}). If we utilize variance-only LayerNorm both on feature embedding and MLP, we observe the similar trend.

As for the influence of variance on MLP, we can see that   large fraction of neuron outputs is pushed into negative number from small positive number after applying variance-only LayerNorm (Figure \ref{fig:fig3}). That means these neuron responses were removed because the following non-linear function is ReLU. We deem this avoid many noises in MLP responses and accelerate the training of the model because of the introduction of the variance.

As for the influence of variance on feature embedding, we can analyze the effect of normalization from another viewpoint. As we all know, the features in CTR tasks are very sparse and there are a large amount of low frequency features. This will lead to the under-fitting of these long tail feature's embedding because there is less training data for them. The under-fitting embedding may contain noise which brings difficulty for feed-forward layer to capture complex feature interactions. We think the correct normalization on feature embedding can alleviate this situation.

We can derive the derivative of $\mathcal{L}$ with respect to input $x_i$ after inserting variance-only LayerNorm as follows. Assume the derivative of $\mathcal{L}$ with respect to $h$ is given, ie. $\frac{\partial\mathcal{L}}{\partial h}$ is known. Then the derivative with respect to input $x_i$ can be write as:

\begin{equation}
\frac{\partial\mathcal{L}}{\partial x_i} = \frac{\partial\mathcal{L}}{\partial h} \ast \frac{\partial h}{\partial x_i}
\end{equation}

\begin{equation}
  \frac{\partial h}{\partial x_i} = \frac{1}{\delta} - \frac{x_i(x_i - \mu)}{\delta^3\cdot H}
  \label{equ:equ10}
\end{equation}
where $\mu$ and $\delta$ are the mean and standard deviation of input. $H$ is the size of input.

We can see that the second sub-item of formula (\ref{equ:equ10}) is nearly zero and can be ignored during the beginning phrase of the model training because parameter initiation approach usually sets the initial value of parameters to be very little random number near zero. So it's the first sub-item that mainly influences the derivative of $h$ respect to input $x_i$. From figure \ref{fig:fig1}, we know that the variance before normalization is rather small positive number. That is to say, the derivative will be made much larger because of the introduction of variance-only LayerNorm. This means the loss will become much more sensitive to the little change of input $x_i$ because of the existence of variance-only LayerNorm.  The variance brings faster convergence for low frequency feature embedding and alleviates the under-fitting of these feature embeddings.

\section{Experimental Results}
In this section, we empirically evaluate the effect of various normalization approaches on deep neural networks on three real-world datasets and conduct detailed studies to answer the following research questions:

\begin{itemize}

\item\noindent\textbf{RQ1} What's the effect of various normalization approaches applied only on feature embedding part of DNN model?

\item\noindent\textbf{RQ2} What's the effect of various normalizations approaches   applied only on MLP part of DNN model?

\item\noindent\textbf{RQ3} What's the effect of various normalization approaches   applied both on feature embedding part and the MLP part of DNN model? Does our proposed variance-only LayerNorm  work?

\item\noindent\textbf{RQ4} Does categorical feature or numerical feature need specific normalization?

\item\noindent\textbf{RQ5} Is there a best normalization combination for DNN model?

\item\noindent\textbf{RQ6} Can we draw the similar conclusion about the effect of normalization in other state-of-the-art models such as DeepFM or xDeepFM?

\end{itemize}

In the following, we will first describe the experimental settings, followed by answering the above research questions.

\subsection{Experiment Setup}
\subsubsection{Datasets}
The following three data sets are used in our experiments:

\begin{enumerate}
  \item \textbf{Criteo\footnote{Criteo \url{http://labs.criteo.com/downloads/download-terabyte-click-logs/}} Dataset:}
  As a very famous public real world display ad dataset with each ad display information and corresponding user click feedback, Criteo data set is widely used in many CTR model evaluation. There are 26 anonymous categorical fields and 13 continuous feature fields in Criteo data set.

  \item \textbf{Malware \footnote{Malware https://www.kaggle.com/c/malware-classification} Dataset:}
    Malware is a dataset from Kaggle competitions published in the Microsoft Malware Classification Challenge. It is almost half a terabyte when uncompressed and consists of disassembly and bytecode malware files representing a mix of 9 different families. All the $82$ feature  fields are categorical.

  \item \textbf{Avazu\footnote{Avazu http://www.kaggle.com/c/avazu-ctr-prediction} Dataset:}
    The Avazu dataset consists of several days of ad click- through data which is ordered chronologically. For each click data, there are 24 fields which indicate elements of a single ad impression.
\end{enumerate}

We randomly split instances by $8:1:1$ for training , validation and test while Table \ref{tab:datasets} lists the statistics of the evaluation datasets. For these datasets, a small improvement in prediction accuracy is regarded as practically significant because it will bring a large increase in a company's revenue if the company has a very large user base.

\begin{table}
\centering
\caption{Statistics of the evaluation datasets}
\begin{tabular}{l|ccc}
\toprule
Datasets  & \#Instances & \#fields & \#features \\
\midrule
Criteo       & 45M  & 39(26 Cat;13 Num) & 30M     \\
Malware     & 8.92M  & 82(all Cat) & 9.89M \\
Avazu       & 40.43M  & 24(all Cat) & 0.64M     \\
\bottomrule
\end{tabular}
\label{tab:datasets}
\end{table}

\subsubsection{Evaluation Metric} 	AUC (Area Under ROC) is used in our experiments as the evaluation metrics. This metric is very popular for binary classification tasks. AUC is equal to the probability that a classifier will rank a randomly chosen positive instance higher than a randomly chosen negative one.  It is insensitive to the classification threshold and the positive ratio. AUC's upper bound is $1$ and larger value indicates a better performance.

\subsubsection{Models for Comparisons}	We mainly use DNN model as the baseline to evaluation the effect of various normalization methods because it's a commonly used component in many current neural network models. DeepFM and xDeepFM are also regarded as another two baselines to further verify the effectiveness of these approaches. Among these baseline models, DNN and DeepFM implicitly capture high order interactions while xDeepFM models high order interactions in explicit way.

\subsubsection{Implementation Details}	We implement all the models with Tensorflow in our experiments. For optimization method, we use the Adam with a mini-batch size of 1000 and a learning rate is set to $0.0001$.  Focusing on normalization approaches in our paper, we make the dimension of field embedding for all models to be a fixed value of $10$. For models with DNN part, the depth of hidden layers is set to $3$, the number of neurons per layer is $400$, all activation function are ReLU. We conduct our experiments with $2$ Tesla $K40$ GPUs.

\begin{table}
  \caption{Overall performance (AUC) of DNN models with various normalization approaches on feature embedding on three datasets}
  \begin{tabular}{l|ccc}
    \toprule
    & Criteo & Avazu & Malware \\
    \midrule
       DNN &
    0.8054 &
    0.7820 &
    0.7263 \\
    \midrule
       +BN &
    0.8066 &
    0.7847 &
    0.7364 \\
       +GN2 &
    0.8096 &
    0.7835 &
    0.7330 \\
       +LN &
    0.8093 &
    0.7848 &
    0.7341 \\
       +S-LN &
    0.8093 &
    0.7843 &
    0.7343 \\
       +VO-LN &
    0.8094 &
    0.7839 &
    0.7358 \\
    \bottomrule
  \end{tabular}
  \label{tab:varnorm_emb_auc}
\end{table}

\subsection{Effectiveness of Normalization on Feature Embedding (RQ1)}
To verify the effectiveness of various normalization approaches on DNN models, comparison experiments are conduced on three evaluation datasets. We add different kinds of normalization either on the embedding layer or hidden layer of a standard DNN model which has 3 MLP layers with 400 neurons per layer. As for GourpNorm, we find GN with 2 groups perform best compared with other group number setting. So we only report the experimental results with this setting (GN2).
We find normalization on feature embedding helps boosting DNN model's performance. The Table \ref{tab:varnorm_emb_auc} shows the experimental results. From Table \ref{tab:varnorm_emb_auc}, we have the following observations:
\begin{enumerate}
  \item If we add normalization on feature embedding, all normalization methods help model's training on three datasets, including BatchNorm, GroupNorm ,LayerNorm, Simple LN and our proposed variance-only LN.
  \item As for some widely used normalization, BatchNorm performs best on Malware dataset while worst on Criteo dataset. That implies the performance of BatchNorm depends on specific datasets and GroupNorm shows the similar trend.  LayerNorm keeps relatively high performance on all three datasets.
  \item As for LayerNorm based normalizations,  LayerNorm,Simple-LN and variance-only LN show comparable performance on all three datasets.
\end{enumerate}

\begin{table}
  \caption{Overall performance (AUC) of DNN models with various normalization approaches on MLP part on three datasets}
  \begin{tabular}{l|ccc}
    \toprule
    &
    Criteo &
    Avazu &
     Malware \\
     \midrule
       DNN &
    0.8054 &
    0.7820 &
    0.7263 \\
    \midrule
       +BN &
    0.8071 &
    0.7836 &
    0.7388 \\
       +GN2 &
    0.8073 &
    0.7836 &
    0.7388 \\
       +LN &
    0.8071 &
    0.7851 &
    0.7378 \\
       +S-LN &
    0.8070 &
    0.7847 &
    0.7376 \\
       +VO-LN &
    0.8075 &
    0.7849 &
    0.7373 \\
\bottomrule
  \end{tabular}
  \label{tab:varnorm_mlp_auc}

\end{table}

\subsection{Effectiveness of Normalization on MLP Part (RQ2)}
We also conduct experiments to apply the different kinds of normalization only on MLP part of DNN model. The overall performances of DNN model with different normalizations on three evaluation datasets are show in the Table \ref{tab:varnorm_mlp_auc}. From the experimental results, we can see that:
\begin{enumerate}
  \item Various normalization approaches show comparable performance on both Criteo and Malware datasets. BatchNorm and GroupNorm slightly underperform LayerNorm based approaches on Avazu dataset.
  \item Compared with normalization only on feature embedding,  normalization only on MLP part performs better on Malware dataset and worse on Criteo dataset on the whole. That may imply that selection between the normalization on embedding and MLP part depends on specific task.
\end{enumerate}

\begin{table}
  \caption{Overall performance (AUC) of DNN models with various normalization approaches on both embedding part and MLP part on three datasets}
  \begin{tabular}{cc|ccc}
    \toprule
    &
    &
    Criteo &
    Avazu &
     Malware \\
     \midrule
         EMB &
       MLP &&&\\
       \midrule
    w/o & w/o & 0.8054 &
    0.7820  &
    0.7263 \\

         +BN &
       +BN &
    0.8068 &
    0.7845 &
    0.7393 \\
        +BN &
       +LN &
    0.8075 &
    0.7863 &
    0.7393 \\
        +BN &
       +VO-LN &
    0.8077 &
    \textbf{0.7869} &
    \textbf{0.7402} \\
        +LN &
       +BN &
    0.8094 &
    0.7838 &
    0.7387 \\
       +LN &
       +LN &
    0.8096 &
    0.7852 &
    0.7372 \\
       +LN  &
       +VO-LN  &
    \textbf{0.8098} &
    0.7857 &
    0.7372 \\
    +VO-LN &
       +BN &
    0.8092 &
    0.7823 &
    0.7394 \\
    +VO-LN &
      +LN  &
    0.8092 &
    0.7841 &
    0.7376 \\
    +VO-LN &
       +VO-LN  &
    0.8097  &
    0.7850 &
    0.7383 \\
    \bottomrule
  \end{tabular}
  \label{tab:varnorm_emb_mlp_auc}
\end{table}

\subsection{Normalization Combination on Both Feature Embedding and MLP (RQ3)}
As discussed in Section 4.2, we can apply the normalization both on the feature embedding part and MLP part of DNN model. Extensive experiments have been conducted and we find the following three normalizations perform better when we combine various normalizations in different part of DNN model: BatchNorm, LayerNorm and Variance-only LN. So we present  experimental results of 9 combinations in Table \ref{tab:varnorm_emb_mlp_auc}.

From the results in Table \ref{tab:varnorm_emb_mlp_auc}, we can see that:
\begin{enumerate}
  \item  If we choose the correct normalization combination, the DNN model performs better than any model which only uses normalization in one part of DNN model, either feature embedding or MLP part. That means they are complementary and it's better to use them both under real-life applications.
  \item  Compared with a standard DNN model, the performances of DNN model with normalizations outperform baseline with a large margin when correct normalization combination are selected.
  \item  If we adopt BatchNorm in normalization combination, the conclusion that its performance depends on dataset still holds.
  \item  Choosing variance-only LayerNorm in MLP part, we usually have relatively higher performance models, no matter which normalization is used in feature embedding part. It tells us that we'd better use VO-LN as  normalization in MLP part when we try to combine the normalizations.
 \end{enumerate}

\subsection{Normalization for Numerical and Categorical Feature (RQ4)}
From the experimental results shown in Table \ref{tab:datasets}, we observe that the performance degrades if we adopt BatchNorm instead of LayerNorm based approaches on feature embedding on Criteo dataset. Considering only the Criteo dataset contains both categorical and numerical feature, we assume that the performance difference is related to the numerical or categorical feature. So we design some normalization combination experiments to testify this assumption. As discussed in Section 4.4, we fix the normalization used in MLP to be variance-only LayerNorm and apply different normalization for numerical and categorical feature. The experimental results can be seen in Table \ref{tab:num_cat_criteo}.

\begin{table}
  \caption{Overall performance (AUC) of DNN models with various normalization approaches for numerical and categorical features on Criteo dataset}
  \begin{tabular}{cc|c|c}
    \toprule
\multicolumn{2}{c|}{EMB} &
MLP &
\\
\midrule
Num &
Cat &
 & \\
\midrule
w/o & w/o & w/o &
0.8054 \\

+BN &
+BN &
+VO-LN &
0.8077\\
+BN &
+LN &
+VO-LN &
0.8068\\
+BN &
+VO-LN &
+VO-LN &
0.8066\\
+LN &
+LN &
+VO-LN &
0.8097\\
+LN &
+BN &
+VO-LN &
\textbf{0.8105}\\
+VO-LN &
+BN &
+VO-LN &
\textbf{0.8107} \\
+VO-LN &
+VO-LN &
+VO-LN &
0.8098 \\
\bottomrule
  \end{tabular}
  \label{tab:num_cat_criteo}
\end{table}

From the results in Table \ref{tab:num_cat_criteo}, we can see that:
\begin{enumerate}
  \item For numerical features, performance of model with LayerNorm or variance-only LayerNorm outperforms the model with BatchNorm. That implies we should utilize LayerNorm based approaches for numerical features.

  \item If we use LayerNorm based normalization for numerical feature and variance-only LayerNorm in MLP, we can see from Table \ref{tab:num_cat_criteo} that we have best performance model with BatchNorm for categorical feature.
\end{enumerate}

\subsection{Performance of NormDNN (RQ5)}
\begin{table}
  \caption{Overall performance (AUC) of DNN model with unified normalization combination( NormDNN) on three datasets}
  \begin{tabular}{l|ccc}
    \toprule
    &
    Criteo &
    Avazu &
     Malware \\
     \midrule
       DNN &
    0.8054 &
    0.7820 &
    0.7263 \\
       DeepFM &
    0.8056 &
    0.7833 &
    0.7295 \\
      xDeepFM &
    0.8063 &
    0.7848 &
    0.7322 \\
       NormDNN &
    \textbf{0.8107} &
    \textbf{0.7869} &
    \textbf{0.7402} \\
    \bottomrule
  \end{tabular}
  \label{tab:normdnn}
\end{table}

If we adopt the following unified normalization combination strategy in DNN ranking model:  variance-only LayerNorm or LayerNorm for numerical feature, BatchNorm for categorical feature and variance-only LayerNorm for MLP, we can gain the best performance model on all three datasets, which achieves significantly better performance than complex model such as xDeepFM.  We call this normalization enhanced model with this unified normalization strategy "NormDNN" in this paper.

The experimental results in Table \ref{tab:normdnn} prove this observation. It is easy to find that NormDNN is more applicable in many  industry applications because of its better performance and high computation efficiency compared with many state-of-the-art complex neural network models.

\begin{table}
  \caption{Overall performance (AUC) of popular models with various normalization approaches on Criteo dataset}
  \begin{tabular}{cc|ccc}
    \toprule
 &
     &
DNN &
DeepFM &
xDeepFM \\
\toprule
     EMB &
   MLP &&&\\
   \midrule
w/o & w/o & 0.8054 &
0.8056 &
0.8063 \\
     +LN &
 w/o &
0.8093 &
0.8100 &
0.8100 \\
  +VO-LN &
  w/o &
0.8094 &
0.8099 &
0.8100 \\
    w/o &
   +LN &
0.8071 &
0.8073 &
0.8075 \\
   w/o &
   +VO-LN &
0.8075 &
0.8076 &
0.8073 \\
+LN &
   +LN &
0.8096 &
0.8099 &
0.8100 \\
+LN &
  +VO-LN &
0.8098 &
0.8102 &
0.8103 \\
+VO-LN &
   +VO-LN &
0.8097 &
0.8101 &
0.8101 \\
    \bottomrule
  \end{tabular}
  \label{tab:varnorm_varmodel_auc}
\end{table}

\subsection{Normalization on DeepFM and xDeepFM Models (RQ6)}
In the following part of the paper, we study the impacts of normalization on two other popular deep neural network models, including DeepFM and xDeepFM. We design some normalization experiments to observe whether it also works for these two models. Notice that the input of FM component in DeepFM is the feature embedding before normalization. The performance of DeepFM degrades if FM component uses the same feature embedding after normalization as DNN component does.

    The results in Table \ref{tab:varnorm_varmodel_auc} show the impact of the various normalizations on model performance. It can be observed that:
    \begin{enumerate}
      \item The performances of both models apparently increase when we add normalization into different parts of model. The experimental results tell us the normalization works for many current state-of-the-art models.

      \item If we select correct normalization combination for simple model such as DNN or DeepFM, the performances of the model with normalization outperform complex model without normalization such as xDeepFM. That means it's more practical to adding normalization on simple models in real-life applications.
    \end{enumerate}

\section{Conclusion}
In this paper, we firstly apply various normalization approaches to the feature embedding part and the MLP part of DNN model.  Extensive experiments are conduct on three real-world datasets and the experiment results demonstrate that the correct normalization significantly enhances model's performance. We also simplify the LayerNorm and propose two new and effective normalization methods in this work. Further more, we find the variance of normalization mainly contributes to this positive effect.

\bibliographystyle{ACM-Reference-Format}
\bibliography{ln}



\end{document}